\documentclass[runningheads]{llncs}

\usepackage{eccv}

\usepackage{eccvabbrv}

\usepackage{graphicx}
\usepackage{booktabs}
\usepackage{multirow}
\usepackage{enumitem}
\usepackage{float}
\usepackage{caption}

\usepackage[accsupp]{axessibility}  % Improves PDF readability for those with disabilities.

\usepackage{hyperref}

\usepackage{orcidlink}

\usepackage{url}
\usepackage{algorithm}
\usepackage{listings}
\usepackage{xcolor}

\definecolor{codecomment}{RGB}{0,120,120}

\begin{document}

% ---------------------------------------------------------------
% TODO REVIEW: Replace with your title
% \title{GarmentZoom: Generating Zoomable Images from Product Photos}
% \title{GarmentZoom: Generating Zoomable Images from Product Photos}
\title{GarmentZoom: Generating Zoomable Images from Garment Listings}

% TODO REVIEW: If the paper title is too long for the running head, you can set
% an abbreviated paper title here. If not, comment out.
% \titlerunning{Abbreviated paper title}

\author{Renjie Zhao\inst{1} \and
Jingwei Ma\inst{1} \and
Huy Huynh Cao\inst{1} \and
Brian Curless\inst{1} \and
Steven M. Seitz\inst{1} \and
Ira Kemelmacher-Shlizerman\inst{1}}

\authorrunning{R.~Zhao et al.}
\institute{University of Washington\\
\email{\{renjiz2,jingweim,huyhuynh,curless,seitz,kemelmi\}@cs.washington.edu}}

% Teaser Image
\onecolumn{
  \maketitle
  \begin{center}
    \includegraphics[width=\textwidth]{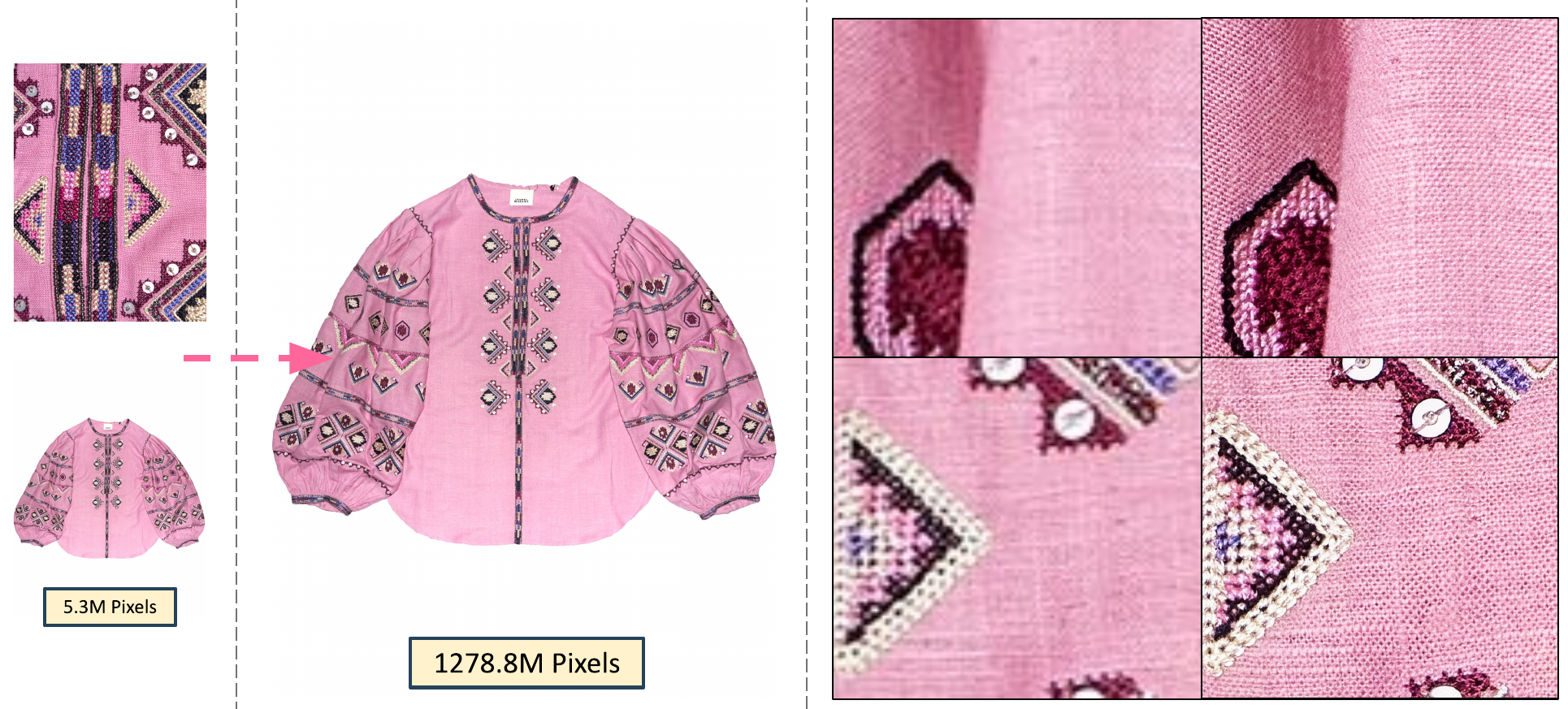}
    \captionof{figure}{Given a standard-resolution full-view garment image and a close-up reference, \textit{GarmentZoom} synthesizes a zoomable high-detail image, enabling seamless exploration for online garment shopping. This specific generated image (middle) upscales the 5-million pixel input by a factor of 12.8$\times$ (linear)  into 1.2 billion pixels. Our method generalizes across diverse garments and supports a continuous scale range of 3--20$\times$. 
    % See the supplemental video for this example and additional full-resolution results.
    }
    \label{fig:teaser}
  \end{center}
  \vspace{-15pt}
}
\begin{abstract}
Online product listings for garments often include an overview photo and a close-up to show garment details.
However, each photo focuses on either field of view or garment detail, forcing users to alternate between views and breaking browsing continuity. 
We present GarmentZoom, a system that enhances the full-view photo to match the fidelity of its accompanying close-up, enabling seamless zoom-and-pan exploration. Unlike standard reference-based super-resolution, our setting involves close-up references that are spatially \emph{unaligned} with the full view, and scale factors that vary substantially across garments (3--20$\times$). Prior work typically relies on alignment to transfer details or requires per-instance fine-tuning to memorize them. 
% Instead, we curate a high-quality apparel dataset from real product listings and train a single model that supports a continuous range of scales across diverse garments. 
Instead, we train a single model that supports a continuous range of scales across diverse garments. 
Our approach synthesizes details without requiring spatial alignment and matches the quality of per-instance methods with a fraction of the training cost. 
See our website for interactive demos of full-resolution results and a patch result gallery at:
\url{https://jason-31.github.io/garmentzoom/}.

\keywords{Texture Synthesis \and Super-Resolution \and Image Generation}
\end{abstract}
\section{Introduction}
\label{sec:intro}

% TODO: check if we need to be more specific about the online shopping problem setting, and introduce the nature of the shopping website images (e.g. clean images)
Online shopping has changed how people experience products, offering convenient visual previews through photos. Yet these photos are limited in resolution: users must alternate between full-views and limited close-ups to examine details, breaking the sense of continuity.  We instead seek to let users inspect any part of a garment at high resolution, as you would be able to do in person.

We introduce \textit{GarmentZoom}, a system that synthesizes a single high-detail garment image from a full-view photo and an \textit{unaligned} close-up. Specifically, our goal is to enhance the full-view photo to the same fidelity as the close-up, using the latter as a reference to guide the generation of garment-specific details. The result is a resolution-enhanced full-view image (Fig.~\ref{fig:teaser}, middle) that enables seamless zoom-and-pan exploration, replacing the fragmented experience of switching between separate views.

Our problem poses distinct challenges not addressed by prior work in general reference-based super-resolution (RefSR). Standard RefSR formulations assume a fixed and discrete upsampling factor (e.g., 4 $\times$). In real product listings, we observe continuous scale factors ranging from 3$\times$ to 20$\times$. Moreover, existing RefSR methods~\cite{zhang2019imagesuperresolutionneuraltexture, yang2020learning, jiang2021robust, cao2022reference} are built around an align-then-transfer paradigm: they establish spatial correspondence between input and reference, then transfer fine-grained details to aligned locations. In our setting, the close-up reference often covers only a small region of the full garment, making spatial alignment infeasible for the rest of the garment. Prior per-instance approaches ~\cite{ma2025ultrazoomgeneratinggigapixelimages} circumvent the variable scale and alignment challenges by fine-tuning on each pair of full-view and close-up images, which is impractical for large-scale deployment.

% To overcome these limitations, GarmentZoom augments a pretrained fixed-scale super-resolution model with a lightweight module that learns to transfer details from the close-up reference. To train this model, we curate a high-quality apparel dataset from an e-commerce site and construct random-scale synthetic low/high-resolution pairs with close-up references. The resulting model matches the quality of per-instance fine-tuning ~\cite{ma2025ultrazoomgeneratinggigapixelimages} while remaining directly applicable to new products without retraining. It supports a continuous range of upscale factors and generalizes to held-out products as well as images from unseen product sites.

% GarmentZoom learns to transfer appearance from the close-up without relying on spatial correspondence, and is trained on a curated apparel dataset with synthetic low/high-resolution pairs across a continuous range of scale factors. The resulting model matches the quality of per-instance fine-tuning~\cite{ma2025ultrazoomgeneratinggigapixelimages} without garment-specific retraining.

To address these challenges, we curate a high-quality apparel dataset from real product listings and construct synthetic low/high-resolution pairs at random scales with close-up references. GarmentZoom learns to transfer appearance from the close-up without relying on spatial correspondence and generalizes to unseen garments and scale factors (3--20$\times$) at test time. The resulting model matches the quality of per-instance fine-tuning ~\cite{ma2025ultrazoomgeneratinggigapixelimages} with low training cost.

\vspace{0.5em}
\noindent Our contributions include:
\begin{itemize}[topsep=3pt,parsep=1ex,leftmargin=*]
    % \item A novel system that converts standard-resolution full-view and close-up product photos into a single high-detail product image, enabling seamless product exploration.
    % \item A lightweight reference-injection module that transfers fine-grained, product-specific texture detail from an unaligned close-up into the low-resolution input.
    % \item A high-quality apparel dataset that contains synthetic low/high-resolution pairs with close-up references, covering diverse apparel products and scale factors.
    \item A system that synthesizes a single high-detail garment image from full-view and close-up product photos, enabling seamless zoom-and-pan exploration.
    \item A reference-guided super-resolution model that supports spatially unaligned close-up references across diverse garments and continuous scale factors.
    \item A data pipeline for constructing LR–HR–reference supervision from real e-commerce product listings.
\end{itemize}
\section{Related Work}
\label{sec:related}

%-------------------------------------------------------------------------
\subsection{Reference-Based Super-Resolution}

Reference-based super-resolution (RefSR) enhances a low-resolution (LR) input by leveraging an additional high-resolution (HR) reference image that provides complementary visual cues. The key challenge lies in establishing reliable correspondences between the LR image and the reference, and transferring appropriate visual details while preserving the LR image's overall structure.

Early approaches construct a database of HR-LR patch pairs synthesized from generic images. At test time, each LR patch is matched to this database using nearest-neighbor retrieval or sparse coding, and the associated HR patch is used to reconstruct the output~\cite{freeman2002example,1315043,5466111}.

Modern RefSR methods leverage deep features for more robust correspondence estimation. SRNTT~\cite{zhang2019imagesuperresolutionneuraltexture} performs patch matching in VGG feature space and injects retrieved textures into the SR network, while TTSR~\cite{yang2020learning} introduces a transformer-based attention module that aggregates reference features through learned query–key similarity. 
And more recently, diffusion-based RefSR methods such as~\cite{wang2026trust} leverage the strong priors of image generation models to achieve enhanced texture synthesis capabilities.

However, these methods still rely on corresponding visual structures of low-resolution (LR) inputs and high-resolution (HR) images. Moreover, these methods assume a fixed super-resolution upsampling factor, making them unsuitable with our task of generating high fidelity details with arbitrary scale between the full-view and close-up images.

A similar recent work, UltraZoom~\cite{ma2025ultrazoomgeneratinggigapixelimages} focuses on real-world LR inputs and HR references with varying upsample scales, but their approach requires hours of fine-tuning per example, which is impractical for our application.

%-------------------------------------------------------------------------
\subsection{Conditioning Generative Models}
\label{subsec:realted-t2i-control}

Advancement in text-to-image generative models has shifted the paradigm of many generative tasks: it is more common to adapt these powerful pretrained priors than training from scratch. Due to the ambiguity of natural language, recent efforts increasingly incorporate image-based controls to enable more precise and reliable guidance.

One popular approach is to employ external modules and inject control signals directly into the model’s hidden states. ControlNet~\cite{zhang2023adding} attaches an auxiliary encoder whose outputs are added to the base U-Net model at multiple resolutions, while LoRA~\cite{hu2022lora} trains lightweight residual weights on selected layers to impose structural or stylistic control.

% Another approach incorporates image conditions through cross-attention. IP-Adapter~\cite{ye2023ip} projects control images into key-value embeddings that are fused with the model’s attention layers, enabling lightweight signal injection without modifying the diffusion backbone. While highly effective for global identity guidance, such methods struggle to provide fine-grained, texture-level control.
Another approach incorporates image conditions through cross-attention. IP-Adapter~\cite{ye2023ip} projects control images into key-value embeddings used by the model’s cross-attention layers, enabling lightweight signal injection without modifying the diffusion backbone. While highly effective for global identity guidance, such methods struggle to provide fine-grained, texture-level control.

\subsection{Texture Synthesis}
\label{subsec:texture-synthesis}
Texture synthesis creates a large, new texture image from exemplar cues by analyzing and reproducing its structural content. Classical non-parametric approaches~\cite{790383, Efros01, kwatra:2003:SIGGRAPH} focus on copying patches from an exemplar. These methods yielding coherent local appearance but often fail to maintain global structural consistency. With recent advancement of deep learning, texture synthesis has become substantially more powerful: CNN-based techniques~\cite{gatys2015texturesynthesisusingconvolutional, ulyanov2016texturenetworksfeedforwardsynthesis, bergmann2017learningtexturemanifoldsperiodic}, GAN-driven models~\cite{zhu2018multimodalimagetoimagetranslation, xian2018texturegancontrollingdeepimage, shaham2019singanlearninggenerativemodel}, and diffusion-based methods~\cite{wang2024infinitetexturetextguidedhigh, zhang2024fabricdiffusion} can generate textures with both high diversity and strong global consistency. Our task similarly synthesizes new texture details from exemplar cues, but operates under the additional constraint of a low-resolution input that dictates the global layout such as fabric deformation, requiring the generated texture to remain consistent with the underlying structure.
\section{Preliminaries}
\label{sec:method-preliminary}

\paragraph{Flux.1-dev ~\cite{flux2024}.}
Our method builds upon Flux.1-dev, a state-of-the-art flow-matching model that learns a continuous velocity field mapping Gaussian noise to images, conditioned on text inputs. Flux.1-dev has a transformer architecture, where each latent image token attends to all other image tokens and to text-conditioning tokens via cross-attention. This mechanism enables the model to integrate global context and semantic guidance during generation.

\paragraph{ControlNet ~\cite{zhang2023adding}.}
% As mentioned in Sec.~\ref{subsec:realted-t2i-control}, 
ControlNet adapts a model for conditional generation by copying selected encoding blocks and fine-tuning the copied weights. At runtime, the control input is encoded by these fine-tuned blocks and added directly to the model's corresponding hidden states during its forward pass. 

\paragraph{LoRA ~\cite{hu2022lora}.}
Low-Rank Adaptation (LoRA) is an efficient parameter-update technique that augments selected linear layers with a low-rank residual. 
For a pretrained weight matrix $W \in \mathbb{R}^{d_{\mathrm{in}} \times d_{\mathrm{out}}}$, LoRA introduces two trainable matrices 
$A \in \mathbb{R}^{r \times d_{out}}$ and $B \in \mathbb{R}^{d_{\mathrm{in}} \times r}$, whose product $BA$ forms a rank-$r$ update while $W$ is kept frozen.
\begin{figure}[!t]
  \centering
    \includegraphics[
      width=1\linewidth,
      keepaspectratio
    ]{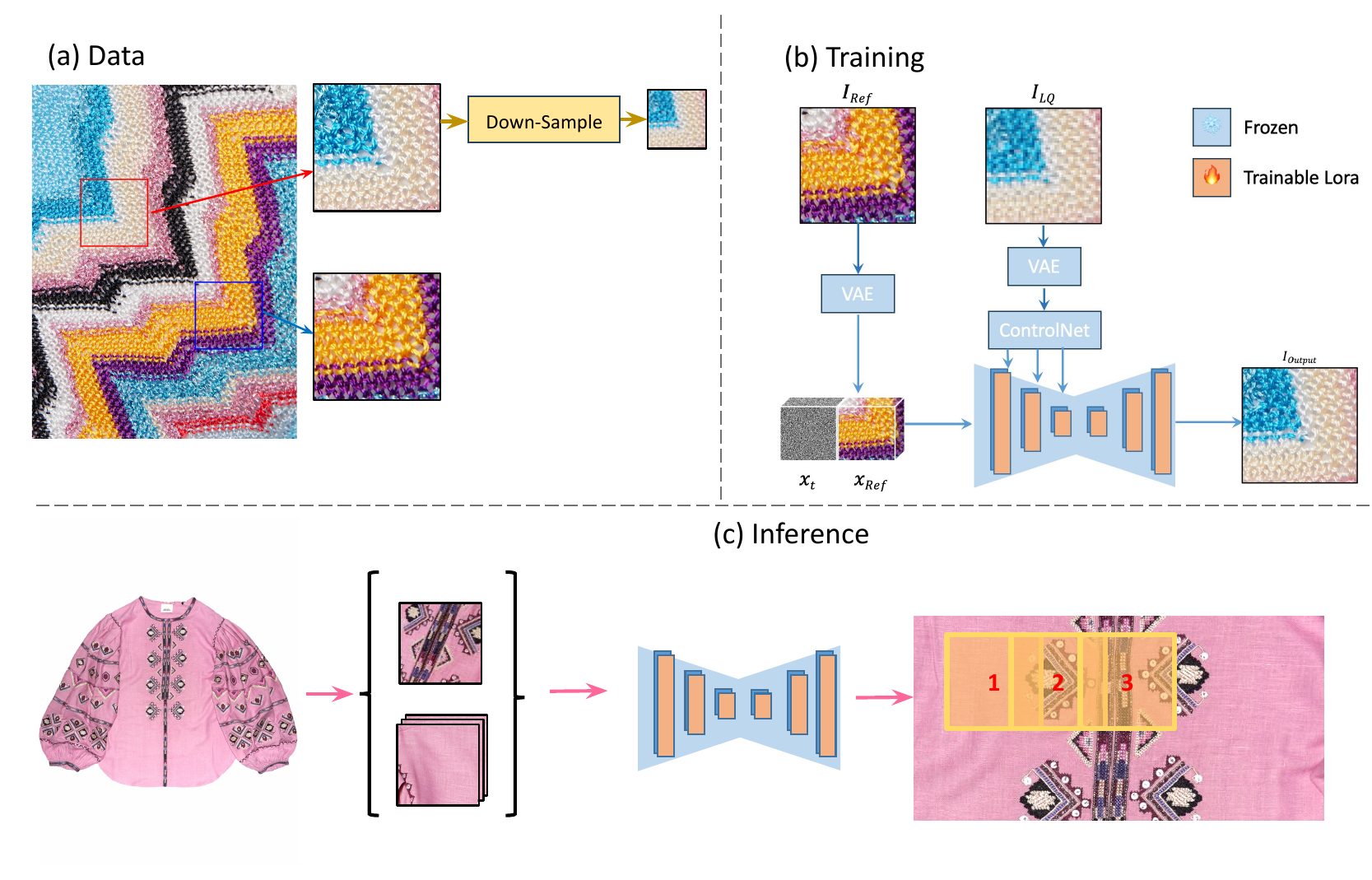}
    \caption{\textbf{Method Overview.}
    (\textbf{a}) \textit{Dataset Construction}: We randomly sample two regions with limited spatial overlap from each high-resolution close-up: a reference $I_{Ref}$ and ground truth $I_{GT}$. The ground truth $I_{GT}$ is downsampled to $I_{LQ}$ to simulate the inference setting. 
    (\textbf{b}) \textit{Training}: We augment the forward pass of a pretrained generative model to adapt it to our garment super-resolution task. We encode the reference $I_{Ref}$ and input $I_{LQ}$ into latent space. A frozen pretrained single-image super-resolution (SISR) ControlNet is conditioned on $x_{LQ}$, while the flow matching model (Flux.1-dev) takes the concatenation of reference latent $x_{Ref}$ and noised latent $x_t$ as input. We fine-tune the model to adapt to our task with LoRA modules. 
    (\textbf{c}) \textit{Inference}: To reduce VRAM usage at inference time, we divide the full-view image into overlapping sliding windows and perform the forward pass on each patch independently. We blend all overlapping patches to produce a coherent result without artifacts.}
    \label{fig:pipeline}
\end{figure}
\section{Method}
\label{sec:method}

Given a pair of standard-resolution photos of a garment product, consisting of a full-view image and a close-up, our goal is to synthesize a single ultra-high-resolution image that preserves the global structure of the full view while incorporating fine-grained details from the close-up. 
% TODO: maybe write this more precisely
% Concretely, we perform reference-guided super-resolution, where the close-up serves as an exemplar to guide detail generation during upscaling.
% Unlike prior approaches that rely on per-instance fine-tuning, we train a single model that generalizes across diverse garment products without requiring product-specific tuning. 
Our system (Fig.~\ref{fig:pipeline}) consists of three stages: dataset construction, model training, and full-resolution inference.

\subsection{Dataset Construction}
\label{subsec:data_prep}

% Product listings typically provide a full-view image that captures global garment structure and at least one close-up image that reveals fine texture details. However, these images are usually captured separately, with different viewpoints, lighting conditions, and camera distances, and the garment may deform between captures. As a result, reliable pixel-level alignment is often infeasible, so we do not train on real full-view/close-up pairs.

% Instead, we synthesize supervised training triplets ($I_{LQ}, I_{Ref}, I_{GT}$) from high-resolution close-up images. For each close-up, we sample two spatially separated crops: one is used as the reference $I_{Ref}$, and the other as the ground-truth target $I_{GT}$. Spatial separation discourages trivial copying and encourages the model to use the reference as the texture cue.

% We obtain the low-quality input $I_{LQ}$ by downsampling $I_{GT}$ with a random continuous scale factor $s \in [3, 20]$, matching the resolution gaps observed in real listings. This yields paired supervision ($I_{LQ}, I_{GT}$) for super-resolution.

% Finally, we use a vision–language model to generate a detailed text description for each garment, which serves as auxiliary conditioning when visual cues are incomplete. We use the same prompt across all patches of the same garment.

Product listings typically provide a full-view image that captures global garment structure and one or more close-up images that reveal fine texture details. However, in practice the close-up usually covers only a small portion of the garment, and even for the overlapping portion, the close-up and the corresponding full-view image region are usually captured with different viewpoints, lighting conditions, and camera distances, and the garment may also deform between captures. As a result, reliable pixel-level alignment is often infeasible, making it difficult to establish real low-/high-resolution pairs from the product photos.

We therefore synthesize supervised training triplets ($I_{LQ}, I_{Ref}, I_{GT}$) from high-resolution close-up images. For each close-up, we sample two spatially separated crops: one is used as the reference $I_{Ref}$, and the other as the ground-truth target $I_{GT}$. Spatial separation discourages trivial solution, encouraging the model to use the reference as a texture cue rather than a source for direct copying. We obtain the low-quality input $I_{LQ}$ by downsampling $I_{GT}$ with a random continuous scale factor $s \in [3, 20]$, matching the resolution gaps observed in real listings. This yields paired supervision ($I_{LQ}, I_{GT}$) for super-resolution.

In addition, we use a vision–language model to generate a detailed text description for each garment, which serves as auxiliary conditioning when visual cues are incomplete. We use the same prompt across all patches of a garment.

\subsection{Training}
\label{subsec:method-training}

\paragraph{Architecture.}
Our method is built on FLUX.1-dev in conjunction with a $4\times$ single-image super-resolution (SISR) ControlNet\footnote{https://huggingface.co/spaces/jasperai/Flux.1-dev-Controlnet-Upscaler}. This base pipeline is trained for fixed-scale super-resolution on general images without reference conditioning. It still provides a strong prior for preserving the underlying low-resolution structure, making it a suitable foundation for reference-guided super-resolution.

% Before we know about OmniControl
% % Our approach to reference-based image guidance augments \textit{FLUX.1-dev}'s forward pass. 
% Our goal is to condition the generation on low-level texture features from a reference image. We achieve this by concatenating a VAE-encoded reference image ($x_{Ref}$) with the noisy latent ($x_t$), which together form the model's new input. To differentiate the conditioning signal from the generation target, we apply distinct timestep embeddings: $T=0$ for $x_{Ref}$ and $T=t$ for $x_t$. We train the model to predict noise only for $x_t$ and discard the noise prediction for $x_{Ref}$ at the end of the forward pass. 
% This effectively enables bidirectional attention between the two streams, which facilitates representing the reference latents in the same latent space as the intermediate latents of $x_t$.
% For efficient fine-tuning, we employ a lightweight LoRA adapter on the flow-matching model, keeping the VAE, text encoders, and ControlNet weights frozen.

To introduce reference-based texture guidance, we utilize a unified-sequence image-conditioning formulation. Specifically, we encode the reference image with the pretrained VAE to obtain a reference latent sequence $x_{\mathrm{Ref}}$, concatenate it with the noisy target latent $x_t$, and process the combined sequence with the FLUX.1-dev transformer. Since the two streams play different roles in our task, we treat the reference latent as an observed conditioning signal and train the model to predict the velocity only for the target latent; the output tokens corresponding to $x_{\mathrm{Ref}}$ are discarded after the forward pass. This allows the target tokens to access reference texture information through the DiT attention layers, while the frozen SISR ControlNet preserves the low-resolution structure.

This unified-sequence conditioning mechanism is closely related to OminiControl~\cite{tan2025ominicontrol}, which processes VAE-encoded image-condition tokens and noisy image tokens jointly in a MMDiT backbone. 
To the best of our knowledge, GarmentZoom is the first work to investigate unified sequence processing in the context of detailed texture conditioning.

\paragraph{Training Objective.}
% Since our objective is to synthesize high-frequency, fine-grained textures, conventional super-resolution losses such as L1, SSIM, or LPIPS are suboptimal, as they encourage over-smoothing and averaged outputs.
We train with the flow matching objective:
\begin{equation}
\mathcal{L}_{\text{FM}} = \mathbb{E}_{x_t, t} \left[ \left\| \hat{u}_\theta(x_t, t, x_{Ref}, x_{LQ}, y) - u(x_t, t) \right\|_2^2 \right]
\end{equation}

\noindent where $x_t$ is the noised latent at timestep $t$, $x_{Ref}$ and $x_{LQ}$ are the encoded reference and low-quality input, $y$ is the text prompt, $\hat{u}_\theta(\cdot)$ is the velocity field parameterized by $\theta$, and $u(x_t, t)$ is the target velocity.

\subsection{Full-Resolution Inference}

Directly generating the output at its full resolution is infeasible due to GPU memory constraints. Therefore, we adopt a sliding-window inference strategy. Following Fig.\ref{fig:pipeline}c, we perform inference on overlapping windows and average the overlapping latent regions on a unified canvas~\cite{bar2023multidiffusion}. However, this method is known to produce artifacts
% ~\cite{wang2024exploiting}
due to the repeated boundaries across inference steps
(discussed in~\cite{wang2024exploiting}). To mitigate this, we introduce sliding windows that vary in position at each inference timestep following ~\cite{ma2025ultrazoomgeneratinggigapixelimages}. This minimizes artifacts from repeated window boundaries, yielding seamless, artifact-free outputs.

\section{Experiments}
\label{sec:experiment}

\subsection{Implementation Details}
\label{subsec:experiment-settings}

\paragraph{Hyperparameters.}
We train our model using LoRA adapters with rank 512 applied to the Flux backbone. We use AdamW with an initial learning rate of $1\times10^{-4}$ and a cosine learning rate schedule with $\eta_{\min}=0$. The total training length is 30k iterations. The effective batch size is 8, implemented using a per-forward batch size of 2 with 4 gradient accumulation steps.

\paragraph{Computational Cost.}
Training our model requires 24 H200 GPU hours. The per-instance SR baseline UltraZoom requires approximately 1.3 hours of A100 GPU time per product. Scaling this to our dataset size (8,547 products) would require over 10,000 GPU hours, since a separate model must be fine-tuned for each product. In comparison, our approach trains a single model that can be directly applied to new products without additional training, resulting in substantially lower overall compute cost.

\paragraph{Dataset.}
Our dataset contains 8,547 high-resolution garment close-up images collected from online product listings. We split the dataset into train, validation, and test sets using a 90/5/5 ratio.
During training, we randomly sample two $512 \times 512$ crops from each close-up image. To better simulate real product imagery, we reject crop pairs with large spatial overlap. These two crops are then used to synthetically construct low-resolution inputs and reference images at random scale factors, following the procedure detailed in Sec.~\ref{subsec:data_prep} and Fig.~\ref{fig:pipeline}b.
% , since the full-view image and close-up references in practice rarely align or overlap due to limited field of view and fabric deformation

% \paragraph{Dataset.}  
% We obtained a total of 8547 closeup images from the internet to be our dataset. We then split the dataset into train, validation and test sets with a ratio of (0.9, 0.05, 0.05).
% During each training step, we randomly extract two 512 $\times$ 512 crops from each closeup image, and reject samples with high overlaps to simulate inference time conditions, where low-resolution inputs and the reference images rarely overlap or align due to the limited field of view of close-ups and fabric deformation. See Fig.\ref{fig:pipeline}b for more detail.
% , we apply spatial constraints during data sampling
% NOTE: Discussion in supplemental?
% See Fig.\ref{fig:pipeline} for detail.
% TODO: check if need. 

% \subsection{Comparison with General RefSR}
\begin{figure}[!t]
    \centering
        \includegraphics[width=1\linewidth]{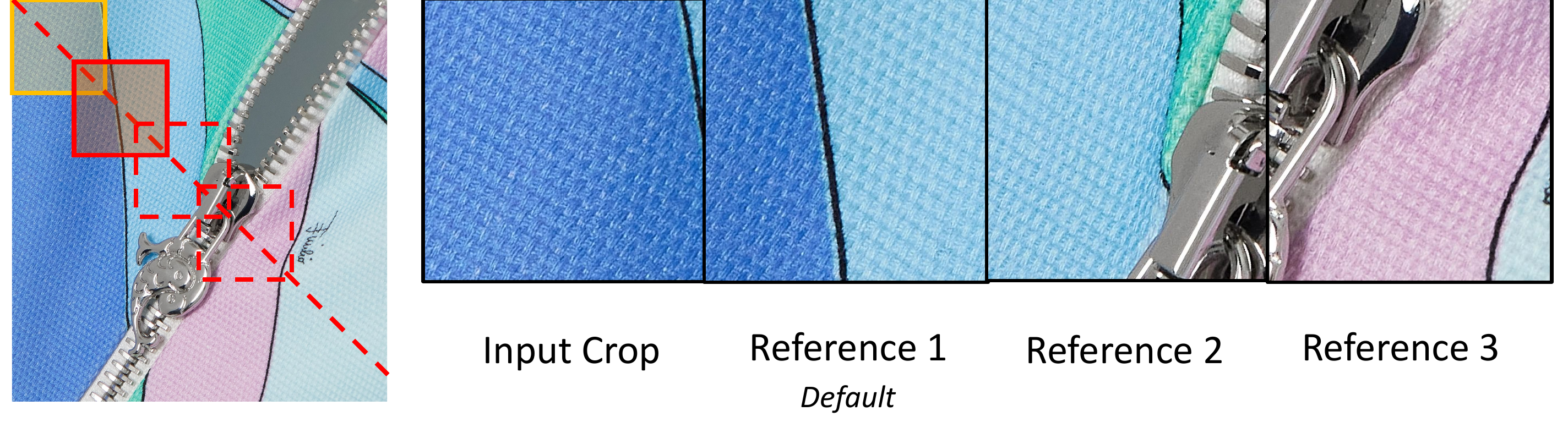}
        \caption{
        % \textbf{Evaluation Crops.} To ensure a fair comparison with baseline method, we fixed the input and reference crop boxes and force a $1\over 9$ overlapping area. Additionally in Sec.\ref{sec:discussion}, to analyze the effect of reference similarity, we move the reference crop box further along the diagonal, and use the distance between input and reference crops as a probe of similarity, resulting reference 2 (small displacement) and 3 (large displacement).
        \textbf{Crop Configuration for Evaluation and Analysis.}
        For baseline comparison, we fix the input and reference crop locations and enforce a $1/9$ overlap between them to ensure consistent evaluation across methods.
        To study the effect of reference similarity (Sec.~\ref{sec:discussion}), we progressively shift the reference crop along the diagonal direction.
        The distance between the input and reference crops acts as a proxy for similarity, producing Reference 2 (small displacement) and Reference 3 (large displacement).
        % We evaluate the effect of reference similarity on reconstruction quality. The low-resolution input crop (yellow box) is held constant, while the reference patch is sampled from along the diagonal (red dashed line). This experiment tests several configurations, from Reference 1 (small overlap), which serves as our default setting for all baseline comparisons, to Reference 3 (large displacement). The quantitative results are shown in Table~\ref{tab:refrence_similarity}, demonstrating higher similarity yields superior results.
        }
        \label{fig:evaluation_crop}
\end{figure}

\subsection{Evaluation Setup}
\label{subsec:experiment-general}

\paragraph{Baseline Methods.}
We compare our method against several RefSR approaches, including AdaRefSR~\cite{wang2026trust}, TTSR\cite{yang2020learning}, DATSR\cite{cao2022reference}, and ReFIR\cite{guo2024refir} (training-free), as well as a Single Image Super Resolution(SISR) method that supports continuous scales, ContinuousSR~\cite{peng2025pixel}. 
For fair comparison, we fine-tune DATSR and AdaRefSR on our curated dataset using their official code and recommended hyperparameters, with random continuous-scale training matching our setup. 
Since DATSR is designed for fixed scaling factors, we train separate models for each reported scale: (4, 10) for Tab.~\ref{tab:sota} and (4, 8, 10, 12, 16) for Fig.~\ref{fig:scale}.

% \paragraph{Baseline Methods.}
% We compare our method against several state-of-the-art RefSR approaches, including TTSR\cite{yang2020learning}, DATSR\cite{cao2022reference} and ReFIR\cite{guo2024refir} (training-free), as well as a SISR method trained with 4--8$\times$ scale, ContinuousSR ~\cite{peng2025pixel}. For fair comparison, we fine-tune DATSR, the most recent trainable RefSR baseline using authors' published code and our data. Due to its architecture constraints, we train individual copy of DATSR at each reported scale (4 and 10 for Tab.~\ref{fig:general-refsr-comparison}, additionally 6,8,12,16 for Tab.~\ref{fig:scale}. 14x ran into an error, as opposed to one model on a conitinuous scale like our setting.

\paragraph{Crop Configurations.}
To comply with the previous methods' assumptions of local feature similarity and ensure fairness, we force overlapping on the synthetic low-resolution and reference pairs to conduct our evaluation. As illustrated in Fig.~\ref{fig:evaluation_crop}, we fix the low-resolution crop to the top-left corner of the full close-up image during evaluation and use reference crop region 1 (small overlap) across all experiments on baseline models. The low-resolution crop is down-sampled using bicubic interpolation to generate synthetic input and ground truth pairs.

\paragraph{Evaluation Metrics.}
For evaluation, we report widely used perceptual metrics including LPIPS and DISTS to measure perceptual similarity between generated and real HR images.
We also report a frequency-domain metric, log-spectral distance (LSD), to capture differences in texture frequency statistics. LSD is computed between texture patches by comparing their radially averaged log power spectra. Given a patch $\mathbf{x}$ and prediction $\hat{\mathbf{x}}$, we compute the power spectrum $P(\mathbf{x}) = |\mathcal{F}(\mathbf{x})|^2$, average it into $K$ radial frequency bins $\bar{P}_{\mathbf{x}}(k)$, and measure the $\ell_2$ distance between the log spectra:
\begin{equation}
\mathrm{LSD}(\mathbf{x}, \hat{\mathbf{x}})
= \sqrt{\frac{1}{K} \sum_{k=1}^{K}
\left(
\log \bar{P}_{\mathbf{x}}(k)
-
\log \bar{P}_{\hat{\mathbf{x}}}(k)
\right)^2 } .
\end{equation}

\noindent Implementation details for LSD are provided in Appendix~\ref{appsec:ablation-lsd}.

\begin{table}[!t]
  \centering
  % \footnotesize
  \resizebox{\linewidth}{!}{%
  \begin{tabular}{@{}lccccc@{\hspace{1.5em}}lccccc@{}}
    \toprule
    \multicolumn{6}{c}{4$\times$} & \multicolumn{6}{c}{10$\times$} \\
    \cmidrule(r){1-6} \cmidrule(l){7-12}
    Method & LPIPS $\downarrow$ & DISTS $\downarrow$ & LSD $\downarrow$ & \multicolumn{2}{c@{\hspace{1.5em}}}{User Study $\uparrow$} &
    Method & LPIPS $\downarrow$ & DISTS $\downarrow$ & LSD $\downarrow$ & \multicolumn{2}{c}{User Study $\uparrow$} \\
    \cmidrule(lr){5-6} \cmidrule(lr){11-12}
    & & & & Qual. & Consis. & & & & & Qual. & Consis. \\
    \midrule
    Input    & 0.285 & 0.234 & 2.028 & {--}  & {--}  & Input    & 0.553 & 0.375 & 2.819 & {--}         & {--}         \\
    TTSR~\cite{yang2020learning}     & 0.137 & 0.135 & 0.846 & {--}  & {--}  & TTSR     & 0.411 & 0.306 & 1.801 & {--}         & {--}         \\
    DATSR~\cite{cao2022reference}   & 0.165 & 0.163 & 1.073 & 13.9\% & 20.8\% & DATSR    & 0.550 & 0.373 & 2.850 & 0.0\%        & 0.0\%        \\
    DATSR-ft & 0.167 & 0.156 & 1.035 & 11.5\% & 16.9\% & DATSR-ft & 0.523 & 0.360 & 2.755 & 2.3\%        & 2.3\%        \\
    AdaRefSR~\cite{wang2026trust} & 0.250 & 0.219 & 0.940 & {--} & {--} & AdaRefSR & 0.322 & 0.264 & 1.145 & {--} & {--} \\
    % AdaRefSR-ft old & 0.222 & 0.232 & 0.807 & {--} & {--} & AdaRefSR-ft old & 0.290 & 0.246 & 1.001 & {--} & {--} \\ %old, wrong, maybe trained not long enough
    AdaRefSR-ft & 0.118 & 0.122 & 0.561 & {--} & {--} & AdaRefSR-ft & 0.205 & 0.175 & 0.857 & {--} & {--} \\
    ContinuousSR~\cite{peng2025continuoussr} & 0.195 & 0.192 & 1.412 & {--} & {--} & ContinuousSR & 0.468 & 0.335 & 2.303 & {--} & {--} \\
    ReFIR~\cite{guo2024refir}    & 0.354 & 0.306 & 1.224 & {--}  & {--}  & ReFIR    & 0.445 & 0.350 & 1.751 & {--}         & {--}         \\
    \midrule
    Ours & \textbf{0.117} & \textbf{0.114} & \textbf{0.455} & \textbf{74.6\%} & \textbf{62.3\%} &
    Ours & \textbf{0.164} & \textbf{0.146} & \textbf{0.512} & \textbf{97.7\%} & \textbf{97.7\%} \\
    \bottomrule
  \end{tabular}}
  \vspace{8pt}
  \caption{
    \textbf{Quantitative comparison with general RefSR baselines and a continuous-scale SISR baseline.} 
    We report results for general RefSR methods (TTSR, DATSR, AdaRefSR, and ReFIR), 
    and a continuous-scale single image super resolution method (ContinuousSR) for $4\times$ and $10\times$ super-resolution. 
    Methods marked with ``-ft'' are fine-tuned on our dataset, while AdaRefSR is also fine tuned with continuous scale to match our setting.  
    Lower is better for LPIPS, DISTS, and LSD. 
    Our method achieves the best perceptual and spectral fidelity across all metrics. User study results (quality and consistency top-1 preference) further support these findings.
}
  \label{tab:sota}
    \vspace{-15pt}
\end{table}

\subsection{Comparison with General Super Resolution Methods}
\begin{figure}[!t]
  \centering
    \includegraphics[width=1.0\linewidth]{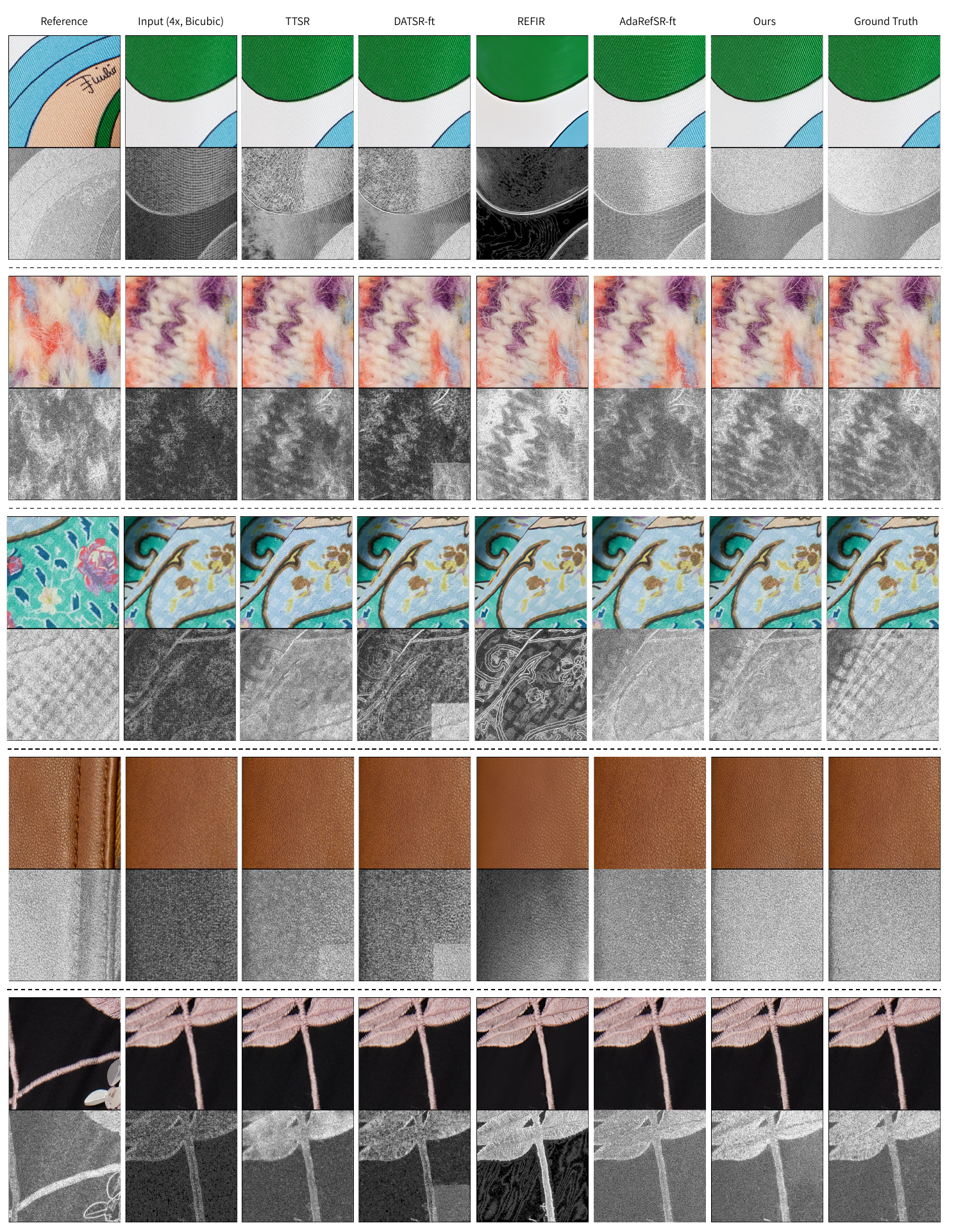}
    \caption{
    \textbf{Qualitative comparison with general RefSR methods}.
    We report results from general RefSR methods (TTSR~\cite{yang2020learning}, DATSR~\cite{cao2022reference}, ReFIR~\cite{guo2024refir}, AdaRefSR~\cite{wang2026trust}) 
    % alongside a continuous-scale SISR method (ContinuousSR~\cite{peng2025continuoussr}) 
    for $4\times$ super-resolution. 
    Methods marked with “-ft” are fine-tuned on our dataset.
    Each garment example spans two rows: the top row shows RGB inputs and results, and the bottom row shows edge maps to better visualize fine structures. From left to right: reference patch (sampled using configuration 1 in Fig.~\ref{fig:evaluation_crop}), bicubic-upsampled input ($4\times$), results from competing methods, our result, and the ground truth. Our method produces textures and structures that are most consistent with the ground truth.
    }
    \label{fig:general-refsr-comparison}
    \vspace{-20pt}
\end{figure}
\paragraph{Quantitative Results.}
Our method shows strong performance on all metrics at both 4× and 10$\times$ scales, shown in Tab.\ref{tab:sota}, 
which delivers the best LPIPS, DISTS, and LSD scores, demonstrating superior texture reconstruction. Moreover, our method shows consistent performance between $4\times$ and $102\times$, while the baseline methods adapt poorly on higher scales.
We further discuss our strength of scale robustness in Sec.\ref{subsec:scale}.

% Our approach outperforms most existing baselines on FID. However, FID primarily captures distribution-level similarity and is less sensitive to localized texture transfer. As we would discuss later, correspondence-based approaches such as DATSR tend to reproduce fine details only within the overlapped area with the reference crop, while non-overlapping areas regress toward the low-resolution input. This behavior would favor the distribution based metric of FID, despite having higher LPIPS and DISTS. In contrast, our method synthesizes coherent, high-detail textures across the entire image, as shown in Fig.~\ref{fig:general-refsr-comparison}.

\paragraph{Qualitative Results.}
We present qualitative comparisons of the general RefSR models in Fig.~\ref{fig:general-refsr-comparison}. All examples run at 4$\times$ scale. At this scale, the low-resolution inputs still retain partial texture information. Both TTSR and DATSR share the limitation where they can only transfer high- fidelity textures in the overlapped areas on many examples, while non-overlapping areas regress to the low-resolution input. 
% This behavior is especially pronounced in DATSR. 
Due to the size limit, such details might be hard to recognize, we also include the edge map 
% obtained by [TODO] 
in the second row of each image to help better visualization. We encourage readers to zoom in to better see the detailed textures of each example.
Despite being optimized for a continuous, dynamic super-resolution scale, our method consistently produces more faithful and spatially coherent textures than the baselines specifically designed for $4\times$ scale.

\begin{figure*}
  \centering
  % Content row
  \begin{minipage}[c]{0.62\linewidth}
    \centering
    \includegraphics[width=\linewidth]{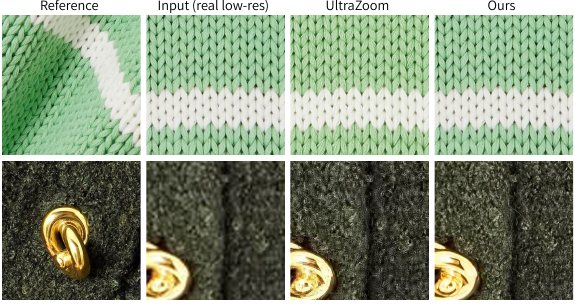}
  \end{minipage}%
  \hfill
  \begin{minipage}[c]{0.35\linewidth}
    \centering
    \resizebox{\linewidth}{!}{%
    \begin{tabular}{lcccc}
      \toprule
      Method & LPIPS $\downarrow$ & DISTS $\downarrow$ & LSD $\downarrow$ \\
      \midrule
      UltraZoom & \textbf{0.185} & 0.152 & \textbf{0.350} \\
      Ours & 0.199 & \textbf{0.128} & 0.586 \\
      \bottomrule
    \end{tabular}}
  \end{minipage}

  \vspace{6pt}

  % Caption row
  \begin{minipage}[t]{0.62\linewidth}
    \vspace{0pt}\setlength{\abovecaptionskip}{0pt}
    \captionof{figure}{
      \textbf{Qualitative Comparison with UltraZoom.} We show two qualitative comparisons with UltraZoom on real low-resolution inputs. The results show that our method are comparable, and sometimes outperforms the per-instance trained method. In the first example, UltraZoom suffers from a color drift, while we maintained color fidelity. In the second example, UltraZoom generated not only the wrong texture, but also generated to the metal part, while our method generate with high texture fidelity and accuracy. 
      }
    \label{fig:ultrazoom-compare}
  \end{minipage}%
  \hfill
  \begin{minipage}[t]{0.35\linewidth}
    \vspace{0pt}\setlength{\abovecaptionskip}{0pt}
    \captionof{table}{\textbf{Quantitative comparison with UltraZoom.} The metrics are measured with synthetic low-resolution input with real downsample factors to match UltraZoom's setting. The results show our method is comparable with per-instance training methods. Bold indicates the better result.}
    \label{tab:ultrazoom-compare}
  \end{minipage}
  \vspace{-15pt}
\end{figure*}

\paragraph{User Study.}
We conduct a controlled user study on visual quality and reference consistency with 13 participants. For each scale ($4\times$, $10\times$), 10 examples were evaluated, yielding 130 responses per scale and criterion (Tab.~\ref{tab:sota}).

\subsection{Comparison with Per-Instance RefSR}
We further compare our method qualitatively with UltraZoom~\cite{ma2025ultrazoomgeneratinggigapixelimages}, a RefSR method with per-instance training. 
Because UltraZoom's per-instance fine-tuning takes 1.5 hours per example, conducting a quantitative evaluation on all the validation instances is impractical. Instead, We  report a quantitative comparison on synthetic low-resolution inputs for 14 examples in Tab.~\ref{tab:ultrazoom-compare}. 

We also provide a qualitative comparison on three examples from the test split, as shown in Fig.~\ref{fig:ultrazoom-compare}. 
% We observe that UltraZoom can overfit to the per-instance training crops, causing the generated colors to drift toward the appearance of the regions used for optimization rather than faithfully matching the input at inference time.
We observe that UltraZoom can exhibit noticeable color drifts and discrepancies in detail, failing to faithfully match the input appearance at inference time.
While UltraZoom benefits from optimizing its weights to one specific product instance at a time, quantitative results show that our generalized model delivers comparable performance, demonstrating strong robustness and cross-product generalization without requiring any instance-specific fine-tuning.

\section{Discussion}
\label{sec:discussion}

\begin{figure*}
  \centering
  % (a) scale_metrics
  \includegraphics[width=1\linewidth]{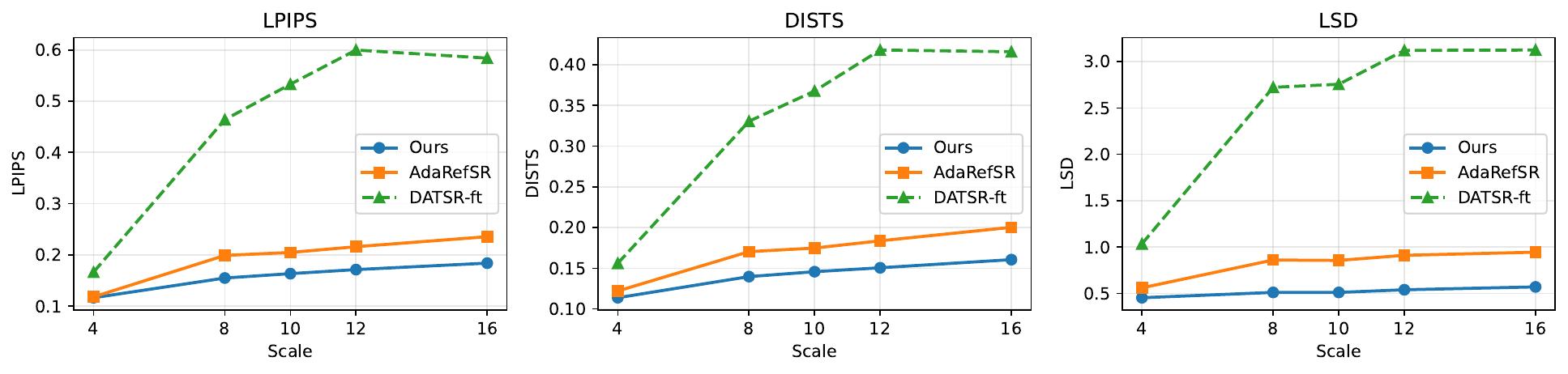}
  \vspace{4pt}
  % (b) multi_scale
  \includegraphics[width=1\linewidth]{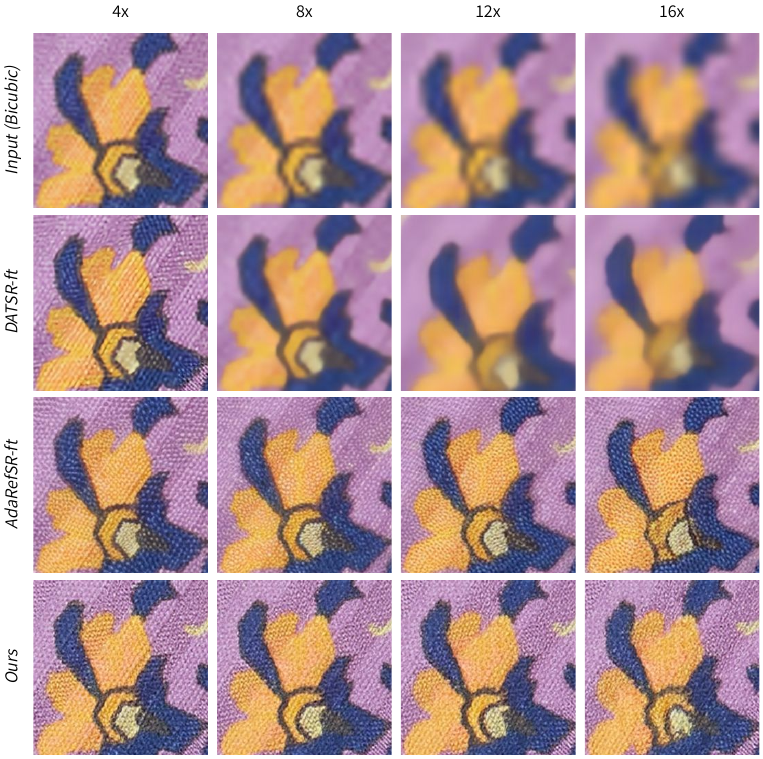}
  \caption{\textbf{Scale robustness.} 
    Scale robustness across large up-sampling factors. We evaluate methods on scales ranging from $4\times$ to $16\times$. \textbf{Top}: Quantitative comparison using LPIPS, DISTS, and LSD. \textbf{Bottom}: Qualitative results for a representative example. While baseline methods degrade rapidly as the scale increases, our method preserves structural details and fabric texture consistently across scales.
% We compare our method with DATSR and ContinuousSR on scales from 4-16.
    % (\textbf{a}) Plot of the quantitative results for each method. 
    % (\textbf{b}) Qualitative comparison of one example against baseline methods.
  }
  \label{fig:scale}
\end{figure*}

% robust against scale
% ===========================================
\subsection{Robustness to Scales}
\label{subsec:scale}
We evaluate our model's robustness to different super-resolution scales. While keeping the input and reference crops the same as in Sec.~\ref{subsec:experiment-general}, we synthesize the low-resolution inputs using different down-sampling scales.

In Fig.~\ref{fig:scale}, we present both the quantitative results plot and a qualitative example.
% Quantitative results across scales are plotted in Fig.~\ref{fig:scale}, and a representative visual comparison is provided in Fig.~\ref{fig:multiscale}. 
Notably, our model maintains strong reconstruction quality even under extreme magnifications of $16\times$, producing both numerical results and visual textures comparable to baseline models at standard 4$\times$ settings. These results indicate that the proposed approach generalizes robustly across a wide range of zoom factors.

% robust against scale
% ===========================================

% \subsection{Effect of Reference Selection}
% \label{subsec:references}
% In this section, we evaluate the effectiveness of different reference-image configurations. As illustrated in Figure~\ref{fig:evaluation_crop}, we consider three reference crops corresponding to (i) small overlap, (ii) small displacement, and (iii) large displacement relative to the target region. Increasing the displacement introduces greater differences in material appearance, color, texture alignment, and fabric deformation, allowing displacement magnitude to serve as a proxy for reference–target similarity. We additionally include a setting in which no reference image is provided. All experiments in this analysis use a fixed downsampling factor of 10$\times$.

% We show quantitative comparison of different reference configurations in Table~\ref{tab:refrence_similarity}. As our model relies on reference image to introduce the texture details, a misaligned reference will prompt our model to generate wrong, while still realistic, textures, we further discuss this in Section~\ref{subsec:limitations}.

\subsection{Effect of Reference Similarity}
\label{subsec:references}
In this section, we study the effect of reference similarity by sliding the reference crop region along the diagonal of the image. 
We consider three reference configurations: (i) small overlap, (ii) small displacement, and (iii) large displacement from the low-resolution input region, as illustrated in Fig.~\ref{fig:evaluation_crop}. 
Larger displacements introduce greater differences in appearance, color, texture alignment, and fabric deformation, and therefore serve as a proxy for increasing reference-target mismatch. To verify this proxy, we additionally measure the cosine similarity between DINOv3 features extracted from the reference crop and the corresponding ground-truth target image. The average DINOv3 similarity decreases monotonically as the displacement increases, from 0.757 under small overlap to 0.643
under small displacement and 0.620 under large displacement, confirming that our spatial displacement protocol induces progressively lower reference-target semantic similarity.
We additionally include a no-reference setting to show the effectiveness of reference images. All experiments in this analysis use a fixed down-sampling factor of 10$\times$.

% In this section, we study the effect of reference similarity by sliding the reference crop region along the diagonal of the image. 
% We consider three reference configurations: (i) small overlap, (ii) small displacement, and (iii) large displacement from the low resolution input region, as illustrated in Fig.~\ref{fig:evaluation_crop}. 
% Larger displacements introduce greater differences in appearance, color, texture alignment, and fabric deformation, and thus serve as a proxy for reference-target similarity. 
% We additionally include a no-reference setting to show the effectiveness of reference images. 
% All experiments in this analysis use a fixed down-sampling factor of $10\times$.

We show quantitative comparison of results with different reference crop regions in Tab.~\ref{tab:refrence_similarity}. As our model relies on reference image to introduce the texture details, a misaligned reference will prompt our model to generate misaligned, while still realistic, textures.

\begin{table*}[t]
\centering
\setlength{\tabcolsep}{4pt} % tighter columns
\renewcommand{\arraystretch}{1.05}

\begin{minipage}[t]{0.39\textwidth}
\vspace{0pt}
\centering
\small
\resizebox{\linewidth}{!}{%
\begin{tabular}{@{}lccc@{}}
\toprule
 & LPIPS $\downarrow$ & DISTS $\downarrow$ & LSD $\downarrow$ \\
\midrule
Input & 0.553 & 0.375 & 2.819 \\
\midrule
Ref \#1, small overlap & \textbf{0.164} & \textbf{0.146} & \textbf{0.512} \\
Ref \#2, small displacement & 0.229 & 0.190 & 0.785 \\
Ref \#3, large displacement & 0.252 & 0.202 & 0.892 \\
\midrule
No Reference & 0.404 & 0.294 & 1.652 \\
\bottomrule
\end{tabular}%
}
\vspace{6pt}
\caption{\textbf{
    Effect of input-reference similarity, evaluated at 10$\times$ scale.} 
    This table corresponds to the visual setup in Fig.~\ref{fig:evaluation_crop}. 
    We test with small overlap, small displacement, large displacement, and No Reference. 
    We also include the low-resolution input before super-resolution.}
\label{tab:refrence_similarity}
\end{minipage}
\hfill
\begin{minipage}[t]{0.55\textwidth}
\vspace{0pt}
\centering
\small
\resizebox{\linewidth}{!}{%
\begin{tabular}{l c c c}
\toprule
Method & LPIPS$\downarrow$ & DISTS$\downarrow$ & LSD$\downarrow$ \\
\midrule
Input & 0.553 & 0.375 & 2.819 \\
Base & 0.470 & 0.310 & 1.515 \\
\midrule
Base + Cross-Attn \#1 (after norm) & 0.360 & 0.273 & 1.199 \\
Base + Cross-Attn \#2 (before norm) & 0.372 & 0.278 & 1.216 \\
\textbf{Base + Ref-Cond LoRA (ours)} & \textbf{0.164} & \textbf{0.146} & \textbf{0.512} \\
\bottomrule
\end{tabular}%
}
\vspace{6pt}
\caption{\textbf{Ablation of our method and alternative conditioning designs.}
All trained variants are trained on scales of 3--20$\times$ and evaluated at $10\times$.
The base model hallucinates generic details, leading to only slight metric improvements.
Our reference-conditioned LoRA substantially improves all metrics.
Cross-attention alternatives improve over the base model but remain worse than our approach.}
\label{tab:ablation}
\end{minipage}
\vspace{-15pt}

\end{table*}

\subsection{Ablation}
We ablate our method and compare our approach against alternative model designs (Tab.~\ref{tab:ablation}). All variants except for Base are trained on our variable-scale dataset (3--20$\times$) and evaluated at 10$\times$. The base model (Flux.1-dev with $4\times$ ControlNet) performs single-image super-resolution without reference conditioning and has not been adapted to the garment domain. It fails to synthesize details given more degraded inputs or hallucinates generic details, resulting in only marginal improvements over the input in all metrics. In contrast, our full method with reference conditioning yields substantial improvements across all metrics, reducing LPIPS and DISTS by more than 50\%.
% and significantly lowering FID.

We further compare against alternative conditioning mechanisms for incorporating reference texture. We experiment with two cross-attention variants inspired by IP-Adapter ~\cite{ye2023ip}, where intermediate latents attend to the VAE-encoded reference image to inject texture information. Despite the conceptual appeal and common use for high-level conditioning (e.g., pose, identity), the cross-attention variants underperform in our setting. In contrast, our design concatenates the reference and input latents at the start of the model, allowing them to co-evolve throughout the transformer blocks and share the same attention space, which enables more effective texture transfer. Additional implementation details and qualitative comparisons are provided in Appendix~\ref{appsec:ablation-implementation}.

\subsection{Attention Visualization}
\label{subsec:attention_visualization}

\begin{figure}[]
\vspace{-15pt}
\centering
\includegraphics[width=0.9\textwidth]{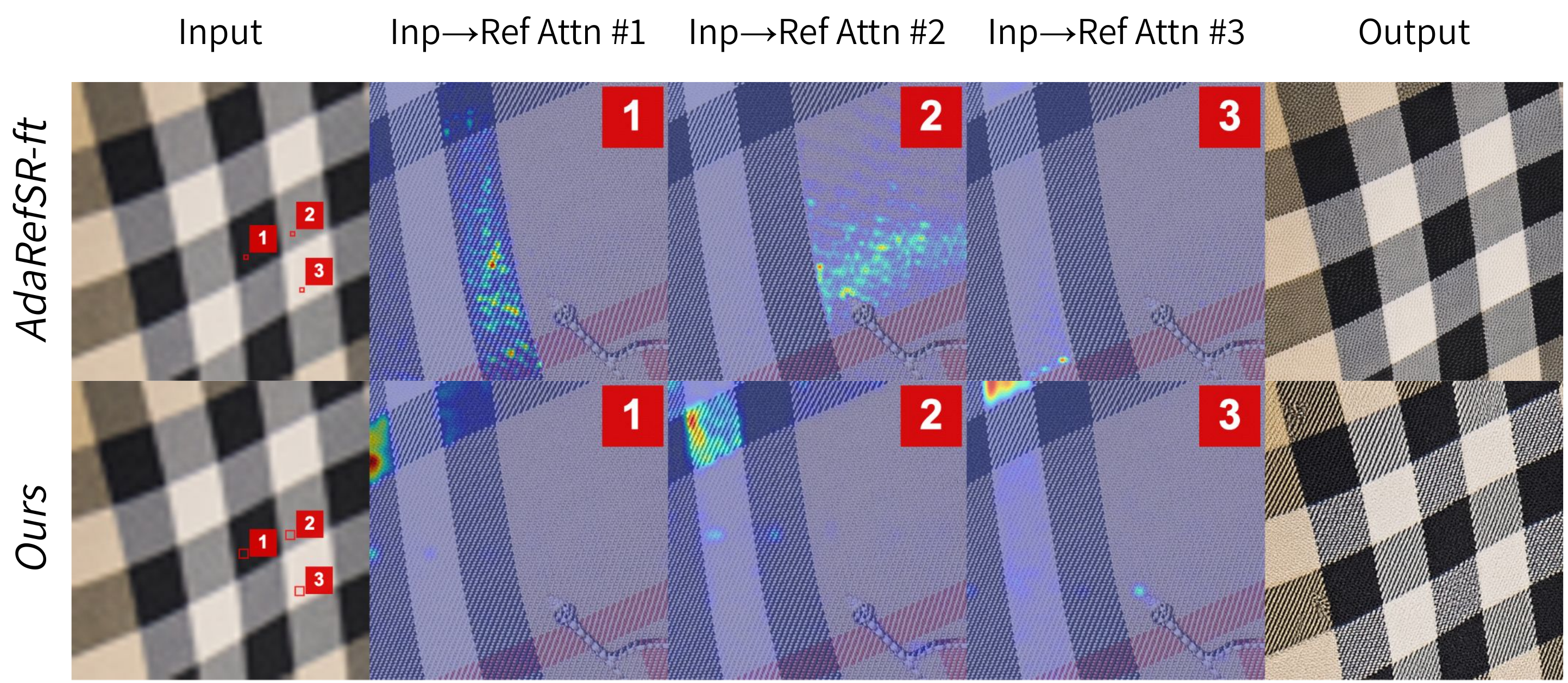}
\caption{
    Attention visualization comparing AdaRefSR (\textit{top}) and our method (\textit{bottom}). 
    \textbf{Column 1}: low-resolution input image with query patches marked in red. 
    \textbf{Columns 2–4}: attention maps showing the reference regions attended by each query patch. 
    \textbf{Column 5}: reconstructed super-resolution result.
}
% \caption{
% Attention visualizations for AdaRefSR (\textit{top}) and ours (\textit{bottom}). Left to right: input with query patches (red), input-to-reference attention heatmaps and outputs.
% }
\vspace{-25pt}
\label{fig:attention_map}
\end{figure}

To show that our method can effectively use the textures from relevant regions of the reference image,  we visualize input-to-reference attention maps in Fig.~\ref{fig:attention_map} and compare our method with fine-tuned AdaRefSR.
AdaRefSR-ft does not consistently attend to reference regions with corresponding
textures, which can lead to incorrect texture transfer. In contrast, our method
focuses more reliably on semantically and texturally related regions in the
reference patch, indicating more robust reference-guided texture transfer.

\subsection{Limitations}
\label{subsec:limitations}
% \input{figures/failure.tex}
% When multiple textures share similar colors and the scaling factor is large, the low-resolution input may lose most cues indicating which texture each region originates from. As a result, our method may confuse textures that exhibit similar color and structure in the degraded input. 
% Fig.~\ref{fig:failure} shows an example where two different white textures appear in the reference, and our method transfers only the top-left texture, likely due to closer color similarity with the input region.
Our method inherits a limitation of the broader RefSR setting: when the required texture is absent from the reference, the model must hallucinate plausible details. A possible direction for future work is to develop a pipeline that automatically selects reference crops that best cover the textures needed for each input region.

\section{Conclusion}
\label{sec:conclusion}
We present GarmentZoom, a system that lifts standard-resolution product images to the fidelity of close-up captures, creating high-detail, zoomable images for seamless product exploration.
Our method extends a pretrained super-resolution model with a lightweight reference-injection module, and trains on synthetic triplets of reference, low-resolution, and high-resolution patches sampled from a high-quality curated dataset of garments.
Our experiments demonstrate that GarmentZoom produces textures consistent with the reference across diverse products and scales, without requiring per-instance tuning or spatial alignment between input and reference.
Overall, the system marks a practical step toward more seamless and engaging virtual product experiences.

% ---- Bibliography ----
%
% BibTeX users should specify bibliography style 'splncs04'.
% References will then be sorted and formatted in the correct style.
\bibliographystyle{splncs04}
\bibliography{main}

% \section*{Acknowledgements}
% Please insert your acknowledgments here.

% ---- Supplementary Material ----

% \newpage
% \renewcommand{\thesection}{S-\arabic{section}}
% \renewcommand{\thefigure}{S-\arabic{figure}}
% \renewcommand{\thetable}{S-\arabic{table}}
% \renewcommand{\theequation}{S-\arabic{equation}}
% \setcounter{section}{0}
% \setcounter{figure}{0}
% \setcounter{table}{0}
% \setcounter{equation}{0}

% \input{sec/9_supplementary}

\clearpage
\appendix

\setcounter{section}{0}

\section*{Appendix}
% \section{Full-Resolution Demos}
% \label{sec:demos}

% To present results at full resolution, we provide two forms of demonstrations in the supplementary directory.

% \paragraph{Recorded video.}
% The supplementary video (\path{recorded_demo.mp4}) contains a screen-recorded walkthrough of full-resolution results across diverse garments and scales, showing continuous zoom-and-pan navigation. The video additionally includes full-resolution side-by-side comparisons with UltraZoom~\cite{ma2025ultrazoomgeneratinggigapixelimages}, a per-instance fine-tuning method.

% \paragraph{Interactive demo.}
% Due to supplementary size limit, we include two interactive examples (\path{interactive_demo/}): one with only our result and the other includes comparison with UltraZoom. The interactive demos allow viewers to explore full-resolution results locally in browser. See \path{README.txt} for viewing instructions.

\section{Alternative Designs}
\label{appsec:ablation-implementation}

\subsection{Implementation Details}
We provide implementation details for the alternative conditioning designs compared in Tab.~4.
Both variants are inspired by IP-Adapter~\cite{ye2023ip-adapter}, which injects reference
conditioning into diffusion models through cross-attention while keeping the backbone frozen
and training only the newly introduced attention projections.
We adapt this idea to MMDiT blocks of FLUX.1-dev.

Unlike IP-Adapter, which encodes the reference image into a high-level CLIP embedding, we instead use the VAE latent $x_{\text{ref}}$ as the input to the key and value projections. Since both $x_{\text{ref}}$ and the noised latent $x_t$ reside in the same latent space, cross-attention between them can facilitate the transfer of low-level texture details.

Formally, queries are computed from the denoising latent, while keys and values are derived
from the reference latent:
\[
Q = x_t W_Q, \quad
K = x_{\text{ref}} W_K, \quad
V = x_{\text{ref}} W_V .
\]
The resulting reference-conditioned update is:
\[
x'_t =
x_t +
\mathrm{softmax}\!\left(
\frac{QK^\top}{\sqrt{d_k}}
\right)V,
\]
where $d_k$ is the key dimension and $W_Q, W_K, W_V$ are the adapter weights.

\begin{figure}
  \centering
  \includegraphics[width=0.8\linewidth]{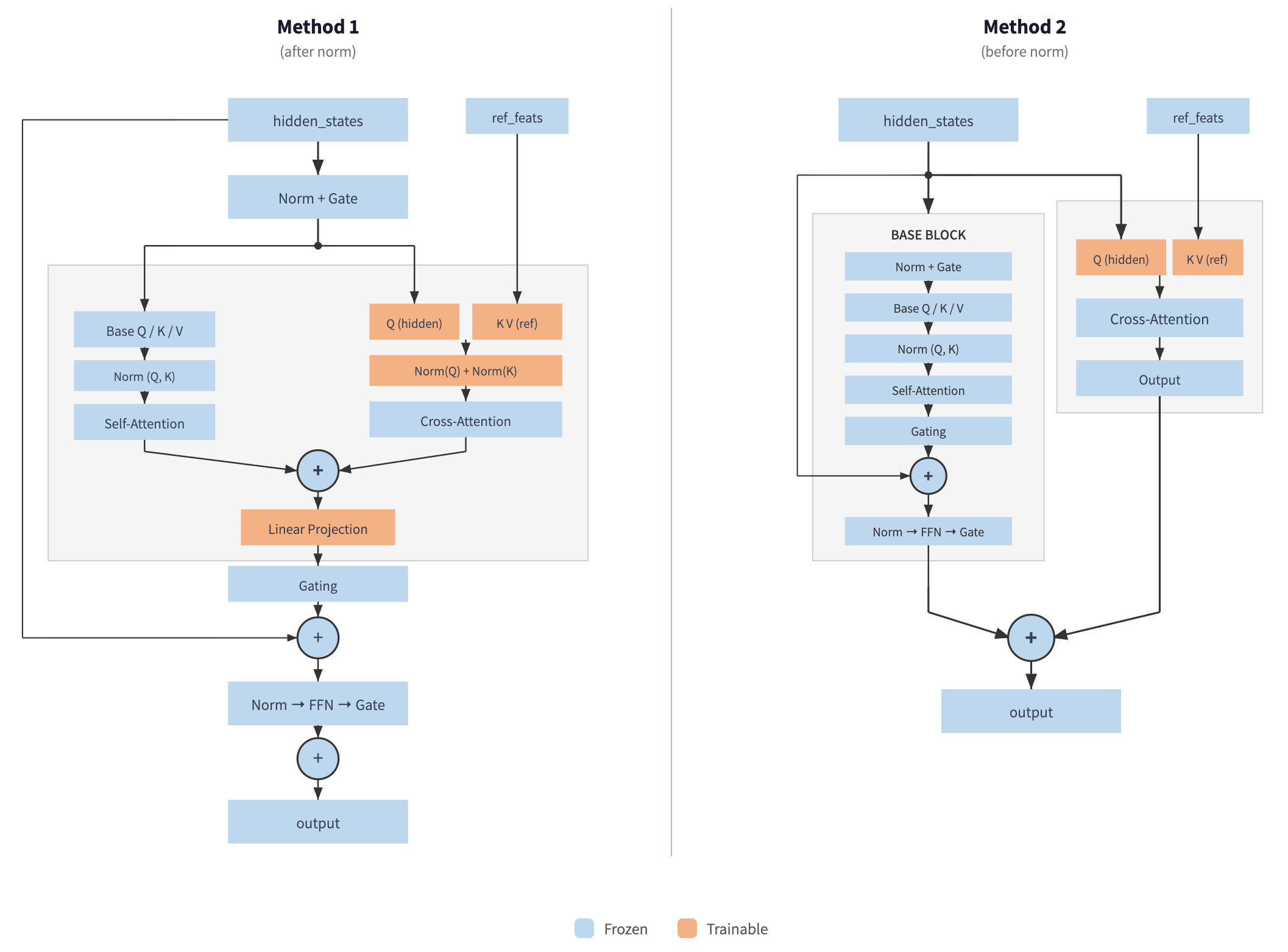}
  \caption{\textbf{Diagram of Ablation Methods Design.} Left (after norm): injects the reference attention within a MMDiT block. Right (before norm): computes the reference attention externally and add it after the MMDiT block.}
  \label{fig:ablation_design}
\end{figure}

The two variants differ in where the adapter attention output is integrated into the
MMDiT block (Fig.~\ref{fig:ablation_design}). The "after norm" variant closely follows the IP-Adapter design and injects the reference
attention within the block's attention stage, allowing the adapter signal to interact
with the block's normalization, residual gating, and feed-forward layers.
The "before norm" variant adopts a more modular design: the query is computed from the block input,
and the reference attention output is added as an additional residual after the entire
MMDiT block, leaving the internal computation of the base block unchanged.

Both variants are less effective than our approach as shown in Tab.~\ref{tab:ablation}, Figs.~\ref{fig:ablation1} and~\ref{fig:ablation2}. We attribute this to the limited
interaction between the reference and denoising representations: cross-attention provides a
one-directional signal where the denoising latent attends to the reference, whereas our
design allows the reference latent to participate fully in the model's bidirectional
attention, enabling richer feature exchange across the network.

\subsection{Qualitative Comparison}
\begin{figure*}[!t]
  \centering
  \includegraphics[width=\textwidth,height=\textheight,keepaspectratio]{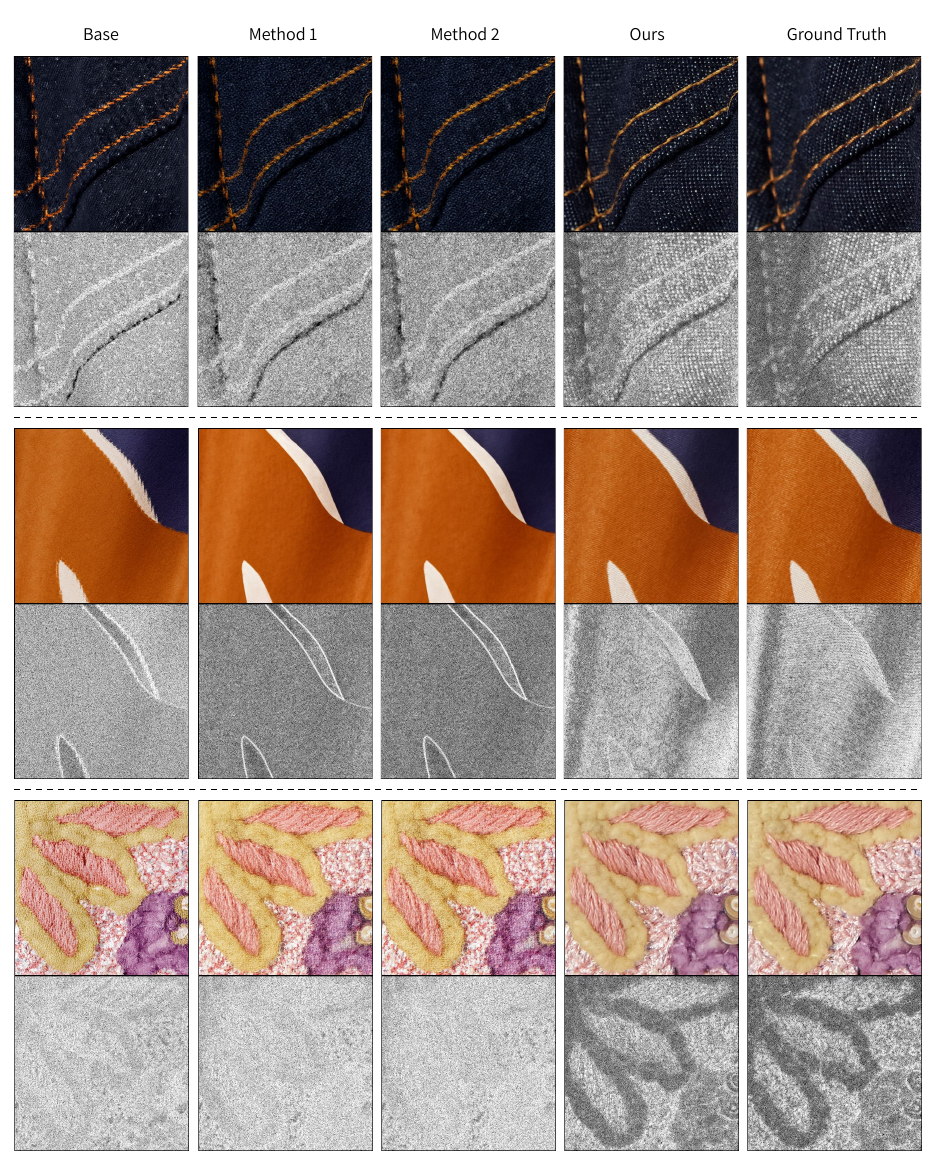}
  \caption{\textbf{Ablation qualitative comparison (10$\times$).}
  Visual comparison of our full method against ablation variants described in Tab.~4 of the main paper. From left to right: input (bicubic upsample), reference patch, Base (FLUX.1-dev $4\times$ ControlNet), Cross-Attn \#1 (after norm), Cross-Attn \#2 (before norm), our full method (Base + Ref-Cond LoRA), and ground truth. Our method produces textures that are most consistent with the reference and ground truth, while the cross-attention variants fail to faithfully transfer fine-grained texture details.}
  \label{fig:ablation1}
\end{figure*}

\begin{figure*}[!t]
  \centering
  \includegraphics[width=\textwidth,height=\textheight,keepaspectratio]{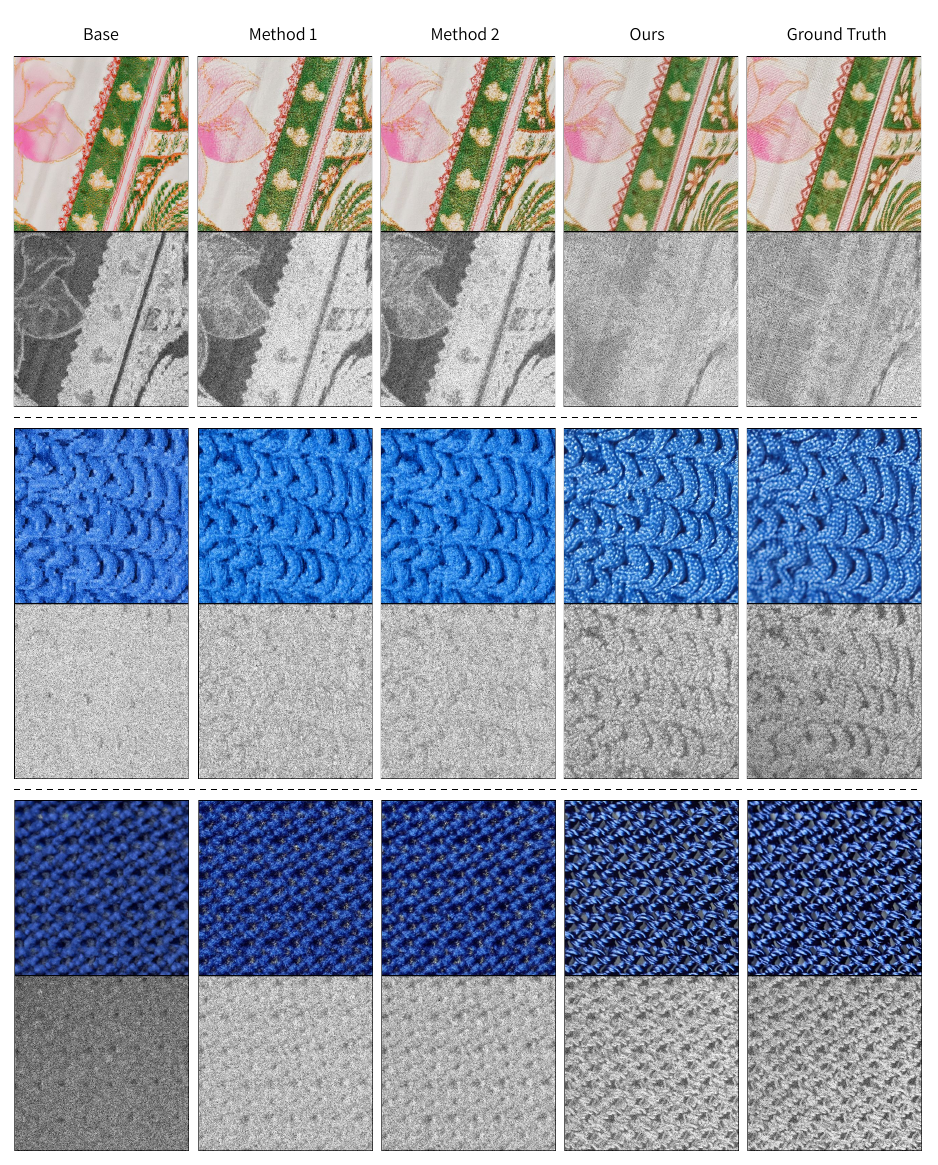}
  \caption{\textbf{Additional ablation qualitative comparison (10$\times$).}
  Additional examples comparing our method against ablation variants. The cross-attention variants tend to either ignore the reference texture or introduce artifacts, whereas our reference-conditioned LoRA consistently synthesizes coherent, high-fidelity textures across diverse garment materials and scales.}
  \label{fig:ablation2}
\end{figure*}

Figs.~\ref{fig:ablation1} and~\ref{fig:ablation2} provide a qualitative view of the
ablation results summarized in Tab.~4 of the main paper. Our backbone model is trained only at $4\times$ and produces noticeable
artifacts when applied at $10\times$. The two cross-attention variants, which introduce
trainable adapter weights, yield more natural-looking outputs with fewer artifacts,
but fail to transfer fine-grained texture details from the reference. Our full method
achieves both: high-quality, artifact-free outputs that reflect the texture
of the reference.

\clearpage

\section{Log-Spectral Distance (LSD)}
\label{appsec:ablation-lsd}

\subsection{Implementation Details}
The LSD metric in Sec.~\ref{subsec:experiment-general} corresponds to the following implementation:
\vspace{-15pt}
\begin{algorithm}[h]
\caption{Implementation of Log-Spectral Distance (LSD)}
\begin{lstlisting}
def _radial_average_power_spectrum(patch):
    f = np.fft.fft2(patch)
    ps = np.abs(np.fft.fftshift(f)) ** 2
    h, w = ps.shape
    cy, cx = h // 2, w // 2
    Y, X = np.ogrid[:h, :w]
    r = np.sqrt((X - cx) ** 2 + (Y - cy) ** 2).astype(int)
    r_max = min(cy, cx)
    raps = np.array([ps[r == i].mean() for i in range(1, r_max + 1)])
    return raps


def calculate_lsd(img1, img2, eps=1e-10):
    if isinstance(img1, Image.Image):
        img1 = np.array(img1)
    if isinstance(img2, Image.Image):
        img2 = np.array(img2)

    if len(img1.shape) == 2:
        img1 = np.stack([img1] * 3, axis=-1)
    if len(img2.shape) == 2:
        img2 = np.stack([img2] * 3, axis=-1)

    if img1.shape != img2.shape:
        raise ValueError(
            f"Images must have the same shape. Got {img1.shape} and {img2.shape}"
        )

    # Convert to grayscale and float
    gray1 = np.mean(img1.astype(np.float64), axis=2)
    gray2 = np.mean(img2.astype(np.float64), axis=2)

    raps1 = _radial_average_power_spectrum(gray1)
    raps2 = _radial_average_power_spectrum(gray2)

    log_raps1 = np.log(raps1 + eps)
    log_raps2 = np.log(raps2 + eps)

    return float(np.sqrt(np.mean((log_raps1 - log_raps2) ** 2)))
\end{lstlisting}
\end{algorithm}
\vspace{-20pt}

\subsection{LSD Metric Validation}
\label{app:lsd_validation}

To validate whether LSD reflects human perceptual preference for garment texture
quality, we reuse the user study samples from Tab.~\ref{tab:sota} and compare each
method's human preference counts with its average LSD on the same samples. As
shown in Tab.~\ref{tab:lsd_ranking}, LSD produces the same method ordering as
human preference at both 4$\times$ and 10$\times$ scales: methods with more human
wins consistently obtain lower LSD. This supports using LSD as a complementary
metric for evaluating texture-frequency fidelity.

\begin{table}[H]
\centering
\footnotesize
\resizebox{\linewidth}{!}{%
\begin{tabular}{lcccc}
\toprule
& \multicolumn{2}{c}{4$\times$} & \multicolumn{2}{c}{10$\times$} \\
\cmidrule(lr){2-3} \cmidrule(lr){4-5}
Method & Human Wins & Avg. LSD$\downarrow$ & Human Wins & Avg. LSD$\downarrow$ \\
\midrule
DATSR    & 50 & 0.748 & 22 & 2.390 \\
DATSR-ft & 37 & 0.836 & 43 & 1.925 \\
Ours     & \textbf{143} & \textbf{0.384} & \textbf{165} & \textbf{0.396} \\
\bottomrule
\end{tabular}}
\vspace{15pt}
\caption{
Validation of LSD against human preference. Lower LSD matches the human
preference ordering at both 4$\times$ and 10$\times$ scales on the same samples
used in the user study.
}
\label{tab:lsd_ranking}
% \vspace{-10pt}
\end{table}

% \section{Attention Visualization}
% \label{app:attention_visualization}

% To show that our method can effectively use the textures from relevant regions of the reference image,  we visualize input-to-reference attention maps in Fig.~\ref{fig:attention_map} and compare our method with fine tuned AdaRefSR.
% AdaRefSR-ft does not consistently attend to reference regions with corresponding
% textures, which can lead to incorrect texture transfer. In contrast, our method
% focuses more reliably on semantically and texturally related regions in the
% reference patch, indicating more robust reference-guided texture transfer.

% \input{figures/supp/attention_map}

\end{document}